\documentclass[runningheads]{llncs}

\usepackage{eccv}

\usepackage{eccvabbrv}

\usepackage{arydshln}
\usepackage{graphicx}
\usepackage{booktabs}

\usepackage{amsmath}
\usepackage{amssymb}
\usepackage{marvosym}
\usepackage{amsfonts}
\usepackage{float}
\usepackage{array}
\usepackage{color}
\usepackage{tabularx}
\usepackage{longtable}
\usepackage{tabu}
\usepackage{multirow}
\usepackage{url}
\usepackage{bm}
\usepackage[noend]{algpseudocode}
\usepackage{algorithmicx,algorithm}
\usepackage{adjustbox}
\usepackage{makecell}
\usepackage{diagbox}

\usepackage[accsupp]{axessibility}  

\usepackage{hyperref}

\usepackage{orcidlink}

\begin{document}

\title{Assessing UHD Image Quality from Aesthetics, Distortions, and Saliency} 

\titlerunning{Assessing UHD Image Quality from Aesthetics, Distortions, and Saliency}

\author{Wei Sun\orcidlink{0000-0001-8162-1949} \and
Weixia Zhang\textsuperscript{\Letter}\orcidlink{0000-0002-3634-2630} \and
Yuqin Cao\orcidlink{0000-0002-5087-6559} \and
Linhan Cao\orcidlink{0009-0007-1871-8757} \and
Jun Jia\orcidlink{0000-0002-5424-4284} \and
Zijian Chen\orcidlink{0000-0002-8502-4110} \and
Zicheng Zhang\orcidlink{0000-0002-7247-7938} \and
Xiongkuo Min\textsuperscript{\Letter}\orcidlink{0000-0001-5693-0416} \and
Guangtao Zhai\textsuperscript{\Letter}\orcidlink{0000-0001-8165-9322}
} 

\authorrunning{Sun \textit{et al.}}

\institute{Shanghai Jiao Tong University \\
\email{\{sunguwei,zwx8981,caoyuqin,caolinhan,jiajun0302,zijian.chen,\\
zzc1998,minxiongkuo,zhaiguangtao\}@sjtu.edu.cn}\\
\textsuperscript{\Letter} Corresponding Authors}

\maketitle

\begin{abstract}
UHD images, typically with resolutions equal to or higher than 4K, pose a significant challenge for efficient image quality assessment (IQA) algorithms, as adopting full-resolution images as inputs leads to overwhelming computational complexity and commonly used pre-processing methods like resizing or cropping may cause substantial loss of detail. To address this problem, we design a multi-branch deep neural network (DNN) to assess the quality of UHD images from three perspectives: \textbf{global aesthetic characteristics, local technical distortions, and salient content perception}. Specifically, aesthetic features are extracted from low-resolution images downsampled from the UHD ones, which lose high-frequency texture information but still preserve the global aesthetics characteristics. Technical distortions are measured using a fragment image composed of mini-patches cropped from UHD images based on the grid mini-patch sampling strategy. The salient content of UHD images is detected and cropped to extract quality-aware features from the salient regions. We adopt the Swin Transformer Tiny as the backbone networks to extract features from these three perspectives. The extracted features are concatenated and regressed into quality scores by a two-layer multi-layer perceptron (MLP) network. We employ the mean square error (MSE) loss to optimize prediction accuracy and the fidelity loss to optimize prediction monotonicity. Experimental results show that the proposed model achieves the best performance on the UHD-IQA dataset while maintaining the lowest computational complexity, demonstrating its effectiveness and efficiency. Moreover, the proposed model won \textbf{first prize in ECCV AIM 2024 UHD-IQA Challenge}. The code is available at \url{https://github.com/sunwei925/UIQA}.
  \keywords{Ultra-high-definition images \and Image quality assessment \and Image aesthetics \and Technical distortions \and Image saliency}
\end{abstract}

\section{Introduction}
\label{sec:intro}

\begin{figure*}[t]
    \centering
    \includegraphics[width=0.6\linewidth]{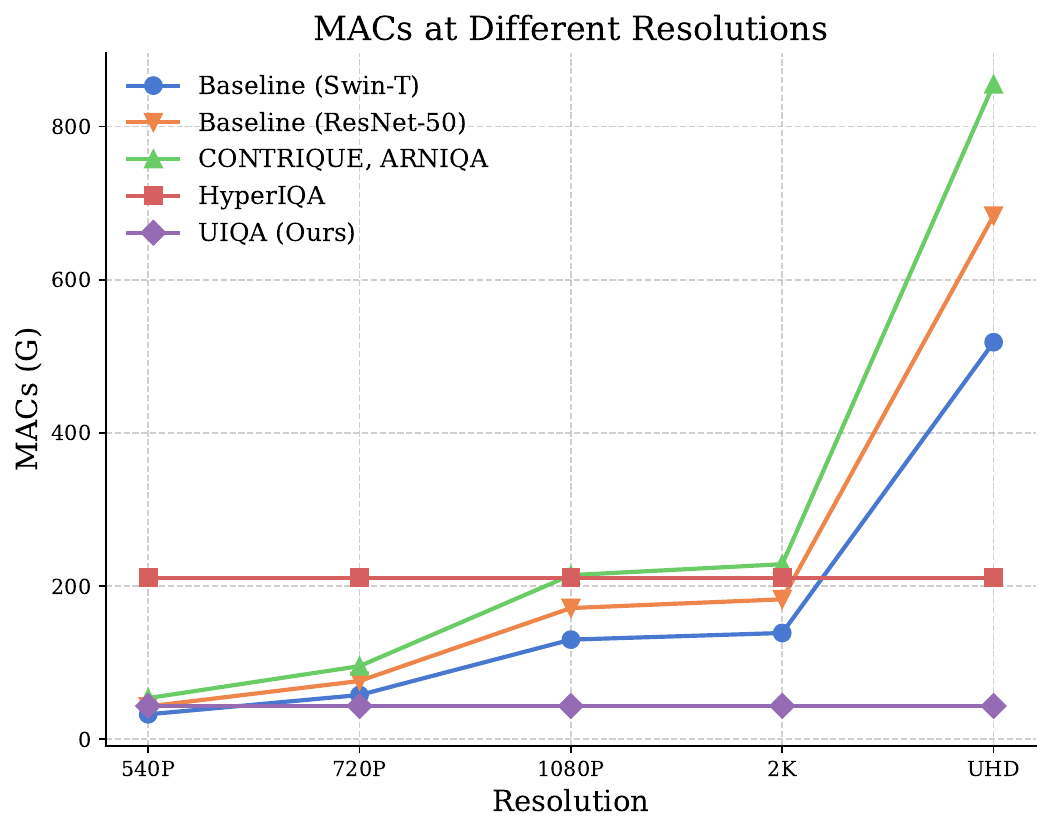}
    \caption{The computational complexity (MACs) of existing IQA methods with different input resolutions.}
    \label{fig:macx}
\end{figure*}

In recent years, ultra-high-definition (UHD) images have been explosively growth on the Internet because of the rapid advancements of computational photography techniques and the popularity of social media networks. Image quality assessment (IQA) is developed to provide quality scores of images aligned to human perception and plays a crucial role in various image processing systems. However, as illustrated in Fig.~\ref{fig:macx}, the extremely high resolutions of UHD images ($4$K or higher) pose a significant challenge to IQA algorithms, as high resolution substantially raises computational complexity, making them impractical for real-time applications and quality monitoring of large-scale UHD images. Thus, it is necessary to develop an efficient and effective IQA model for UHD images.

Existing IQA methods~\cite{zhai2020perceptual} can be broadly classified into two categories: handcrafted-based methods~\cite{moorthy2011blind, saad2012blind, mittal2012no, mittal2012making, zhai2019free, liu2019unsupervised,zhai2021perceptual,zhang2021no, zhang2022no,zhang2021full,li2022full,min2016blind,min2018blind} and deep neural network (DNN)-based methods~\cite{kang2014convolutional, bosse2017deep, ma2017end, zhang2020blind, su2020blindly, ke2021musiq , zhang2021uncertainty, sun2023blind, zhang2023blind, madhusudana2022image, saha2023re, zhao2023quality, agnolucci2024arniqa, sun2024dual}. Handcrafted-based IQA methods primarily leverage prior knowledge of visual quality perception to design quality-related features to quantify image distortions. Popular handcrafted features include natural scene statistics (NSS) features~\cite{moorthy2011blind, saad2012blind, mittal2012no, mittal2012making}, free energy features~\cite{zhai2019free,zhai2011psychovisual,gu2014using}, image texture~\cite{liu2019unsupervised,zhai2021perceptual,zhang2021no}, corner/edge~\cite{zhang2022no,zhang2021full,li2022full,min2016blind,min2018blind}, etc. Despite their utility, handcrafted features are limited in modeling the impact of varying content on visual quality. Hence, they perform well on synthetically distorted images, where quality degradation is mainly due to synthetic distortions, but perform poorly on authentically distorted images, where image quality is influenced by both image content and in-the-wild distortions~\cite{sun2023blind}. DNN-based IQA methods~\cite{kang2014convolutional, bosse2017deep, sun2023blind,zhang2023blind,madhusudana2022image,saha2023re,zhao2023quality,agnolucci2024arniqa} automatically learn quality-aware features by training a carefully designed deep neural network (DNN) in a learning-based manner, which have demonstrated strong performance in evaluating the quality of both synthetically and authentically distorted images. However, as shown in Fig.~\ref{fig:macx}, the computational complexity of DNN increases dramatically with the resolution of the input image, leading to extremely high computational demands when using full-resolution UHD images as inputs. Common image pre-processing techniques for DNN, such as downsampling and cropping, while significantly decreasing the computations, make the UHD image lose substantial details or lack content completeness, thus compromising the effectiveness of UHD image quality evaluation. Moreover, existing studies~\cite{hosu2020koniq,sun2024analysis} indicate that it is difficult to optimize DNN-based IQA methods with high-resolution inputs.

\begin{figure*}[t]
    \centering
    \includegraphics[width=1\linewidth]{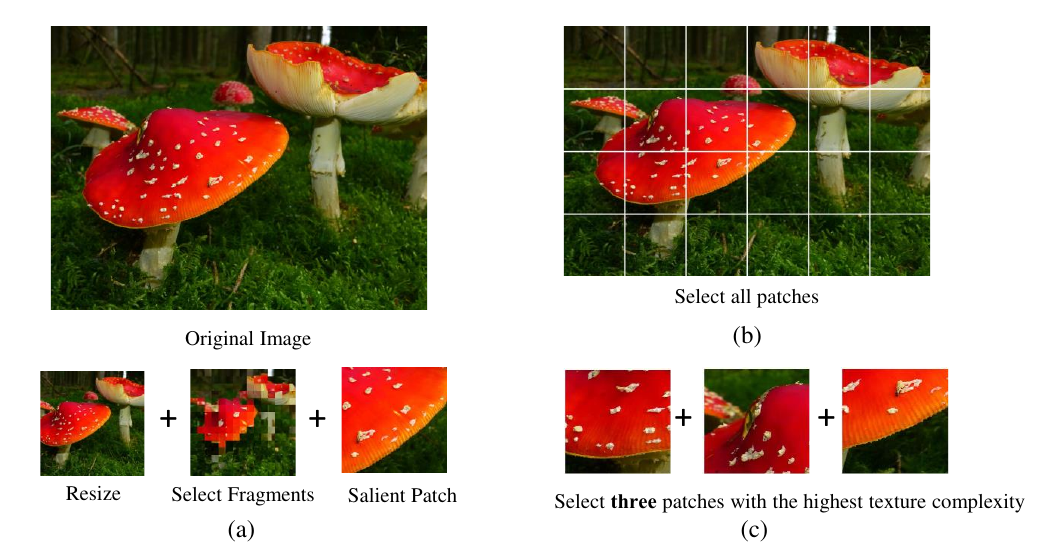}
    \caption{The different image pre-processing methods for UHD images. (a) is the proposed method, which utilizes the resized image, the fragment image, and the salient patch to extract features of aesthetic, distortion, and salient content. (b) samples all non-overlapped image patches for feature extraction~\cite{tan2024highly}. (c) selects three representative patches with the highest texture complexity for feature extraction~\cite{zhu2021perceptual,lu2022deep}.}
    \label{fig:pre-processing}
\end{figure*}

In general, there are two primarily factors that affect human perception on image quality: aesthetic characteristics~\cite{kao2017deep,wang2022deep} and technical distortions~\cite{wu2022fast,sun2023blind,sun2022deep,sun2024enhancing}. The aesthetic characteristics refer to human preferences for image content, color style, layout, etc., which are typically global image features and resolution-insensitive. In contrast, the technical distortions refer to image degradation types such as blur, noise, contrast, compression, etc., which are noticeable in local patches and resolution-sensitive. In addition, image saliency~\cite{borji2012state} is also crucial for image quality assessment, particularly for UHD images. That is because UHD images are more suitable for display on large screens, where human attention is more easily attracted to the salient regions of images. Thus, the perceived quality of UHD images is closely related to the visual quality of these salient regions of images~\cite{zhang2014vsi,yang2019sgdnet}. 

By individually evaluating the quality of these dimensions, we can effectively reduce computational complexity using appropriate preprocessing techniques. Specifically, for aesthetic quality evaluation, we can aggressively downsample the UHD image into a low-resolution one, which loses textural detail but still preserves the aesthetic-related features. For technical distortion measurement, we can extract local mini-patches from the original image, which compromises the completeness of semantic content but still effectively captures technical distortions. For salient content perception, we can use the saliency detection algorithm to crop the most salient region of the UHD image, representing the most critical content of the UHD image for human visual perception. As a result, we can use several low-resolution images instead of the full-resolution images for quality-aware feature extraction, thus achieving significantly lower computational complexity. We illustrate this process in Fig.~\ref{fig:pre-processing} (a).

Therefore, we propose an efficient UHD IQA model that evaluates the perceptual quality of UHD images from three perspectives: \textbf{global aesthetic characteristics}, \textbf{local technical distortions}, and \textbf{salient content perception}. As depicted in Fig.~\ref{fig:model_diagram}, our model comprises three branches, each dedicated to one quality dimension. For the aesthetic branch, we resize and crop the original UHD image into a resolution of $480\times480$ for aesthetic analysis. For the distortion branch, we adopt the grid mini-patch sampling method~\cite{wu2022fast} to uniformly sample local patches across the entire image. Specifically, it first divides the UHD image into $15\times15$ equal-sized patches and then randomly crops a mini-patch with the resolution of $32\times32$ from each patch. These mini-patches are further spliced into a composite image with a resolution of $480\times480$ for distortion analysis. For the branch of salient content perception, although the optimal approach is to utilize a state-of-the-art saliency detection algorithm to identify the most salient region, it would significantly increase computational complexity. Hence, we opt for a simpler approach by exploiting the center bias effect in saliency detection~\cite{borji2012state}, which crops the center patch with a resolution of $480\times480$ from the UHD image to assess salient content quality. To keep the model simplistic, we utilize the standard Swin Transformer Tiny model as the feature extraction networks. We pre-trained the feature extraction network on AVA~\cite{murray2012ava}, a large-scale aesthetic image assessment dataset containing $250,000$ images, to enhance the networks' capability for quality-aware feature analysis. We employ both the mean square error (MSE) loss and the fidelity loss~\cite{tsai2007frank} to optimize the proposed model. The MSE loss specifically reduces the absolute error relative to the ground truth, whereas the fidelity loss improves the rank consistency between two image pairs. The experimental results on the UHD IQA dataset~\cite{hosu2024uhd} demonstrate that our model can effectively assess the quality of UHD images with acceptable computational complexity.

In summary, the contributions of this paper are concluded as follows:
\begin{itemize}
\item We propose an efficient and effective IQA model to evaluate the quality of UHD images from perspectives of global aesthetic characteristics, local technical distortions, and salient content perception.
\item We aggressively downsample UHD images into three low-resolution ones for corresponding dimension feature extraction, thus significantly reducing the computational complexity of the proposed model.
\item The experimental results show that the proposed model achieves the best performance on the UHD IQA dataset, even with a simplistic feature extraction network, demonstrating its effectiveness and its broad applicability in real-world scenarios.
\end{itemize}

\section{Related Work}
\label{sec:related_work}

\subsection{General-purpose IQA Methods}

\noindent\textbf{Handcrafted-based IQA methods.} Natural scene statistics (NSS) is the most commonly used handcrafted features for IQA. The motivation behind NSS-based IQA methods is that high-quality images adhere to specific statistical properties, whereas quality degradations cause deviations from these statistics. For example, DIIVINE~\cite{moorthy2011blind} employs a generalized Gaussian distribution to capture NSS features in the wavelet domain, which are subsequently used to identify the distortion types and regress distortion-specific quality scores. BLIINDS-II~\cite{saad2012blind} and BRISQUE~\cite{mittal2012no} further extract NSS features in the discrete cosine transform (DCT) domain and the spatial domain, respectively, and directly regress the extracted features into quality scores. NIQE~\cite{mittal2012making} measures the distance between NSS features of distorted images and those of a set of high-quality images as quality scores. In addition to NSS, other handcrafted features such as free energy \cite{zhai2019free,zhai2011psychovisual,gu2014using}, image texture~\cite{liu2019unsupervised,zhai2021perceptual,zhang2021no}, corner/edge~\cite{zhang2022no,zhang2021full,li2022full,min2016blind,min2018blind}, etc. are also used in previous IQA models. 
Nevertheless, handcrafted-based IQA methods are usually designed for synthetically distorted images and perform poorly on in-the-wild images.

\noindent\textbf{DNN-based IQA methods.} 
A typical DNN-based IQA model consists of a feature extraction network to extract quality-aware features and a regression network to map the features into quality scores. For example,
Kang \textit{et al.}~\cite{kang2014convolutional} use a shallow CNN model to estimate the patch scores and the image-level quality score is averaged by the evaluated patch scores of the corresponding image.
Bosse \textit{et al.} \cite{bosse2017deep} further deepen the CNNs model by jointly learning the quality and weight of patches.
Zhang \textit{et al.} \cite{zhang2020blind} utilize two kinds of CNN models to obtain a better representation, where two CNN are respectively pre-trained on the distortion type and level classification task and the image classification task.
Su \textit{et al.} \cite{su2020blindly} develop a self-adaptive hyper network to aggregate local distortion features and global semantic features.
Ke \textit{et al.} \cite{ke2021musiq} propose a multi-scale image quality Transformer (MUSIQ), which utilizes the transformer architecture to solve problems of images with varying sizes and aspect ratios.
Sun \textit{et al.}~\cite{sun2023blind} propose a staircase network to extract features from intermediate layers of CNN to enhance the quality-aware feature representation.
Chen \textit{et al.}~\cite{chen2024topiq} propose a top-down approach that uses high-level semantic information to guide the IQA network to focus on semantically important local distortion regions.

Generalization is a concern for DNN-based IQA methods. To enhance the generalization of IQA models, some studies~\cite{zhang2021uncertainty,sun2023blind,zhang2023blind} train feature extraction networks on multiple datasets to increase the diversity of training samples. Unsupervised learning methods~\cite{madhusudana2022image,saha2023re,zhao2023quality,agnolucci2024arniqa}, such as contrastive learning, are also utilized to train feature extraction networks on large-scale unlabelled images. Recently, due to the powerful zero-shot learning ability of visual-language models, such as CLIP, large multi-modality models (LMMs), some studies~\cite{wang2023exploring,agnolucci2024quality,zhang2023blind,wu2023q} fine-tune or directly inference visual-language models with the quality-related textual prompts for IQA tasks.

\subsection{High-resolution IQA Methods}
Existing high-resolution IQA methods primarily address two issues. First, with the rapid advancement of super-resolution techniques, there exists a significant number of fake UHD images/videos on the Internet. To improve the quality of experience (QoE) of users, it is essential to assess whether UHD images/videos were captured by cameras or upsampled by super-resolution methods, as the latter consume high transmission bandwidth but deliver a subpar QoE. Rishi \textit{et al.}~\cite{shah2021real} propose a two-stage approach to determine whether a video frame has real or fake 4K resolution. In the first stage, local patches are classified using a lightweight CNN, and in the second stage, these local assessments are aggregated into a global image-level decision using logistical regression. However, this approach cannot evaluate the quality scores of 4K images, which limits its applicability. Zhu \textit{et al.}~\cite{zhu2021perceptual} further develop an IQA metric to distinguish true and pseudo 4K images and simultaneously measure their quality. It extracts frequency domain features and NSS-based features from three representative patches with the highest texture complexity and then maps these features into the true/false 4K labels and the quality scores. Lu \textit{et al.}~\cite{lu2022deep} propose a grey-level co-occurrence matrix (GLCM) based texture complexity measure to select three representative image patches from a 4K image, which are subsequently utilized to extract quality-aware features from the intermediate layers of the convolutional neural network (CNN). The extracted features are regressed into the class probability and the quality score by two multilayer perceptron (MLP) networks via the multi-task learning manner.

Second, with the growing number of UHD images on social media platforms, it is also crucial to assess the quality of in-the-wild UHD images, which on the one hand can optimize the image processing algorithms in camera imaging systems, on the other hand, can help media providers further compress and transmit UHD images efficiently. Huang \textit{et al.}~\cite{huang2024high} construct a high-resolution image quality database (HRIQ), consisting of $1,200$ images with a resolution of $2880\times2160$. They further propose a HR-IQA method~\cite{korhonen2021consumer}, which employs a CNN to extract features from image patches, followed by training a recurrent neural network separately to estimate the quality scores. Hosu \textit{et al.}~\cite{hosu2024uhd} also build a large-scale UHD-IQA dataset containing $6,073$ 4K images with a fixed width of $3840$ pixels, focusing on highly aesthetic photos of high technical quality. Tan \textit{et al.}~\cite{tan2024highly} propose a full-pixel covering (FuPiC) training strategy for 4K video quality assessment. They divide the full-resolution video into equal-sized patches, integrate frequency information into these patches, and then obtain patch-level scores and weights using a video Swin Transformer. Finally, the overall video quality score is computed as a weighted sum of the patch-level quality scores.

In summary, UHD image authentic assessment studies typically extract features from a few selected high-frequency patches, which results in low computational complexity but fails to capture the overall image quality characteristics. On the other hand, in-the-wild UHD IQA studies primarily divide the images into dense image patches to extract features of the entire image content, which are integrated and regressed into quality scores. However, the large number of patches inevitably leads to high computational complexity. We illustrate these two pre-processing Fig.~\ref{fig:pre-processing} (b) and (c).

\section{Methods}
\begin{figure*}[t]
    \centering
    \includegraphics[width=1\linewidth]{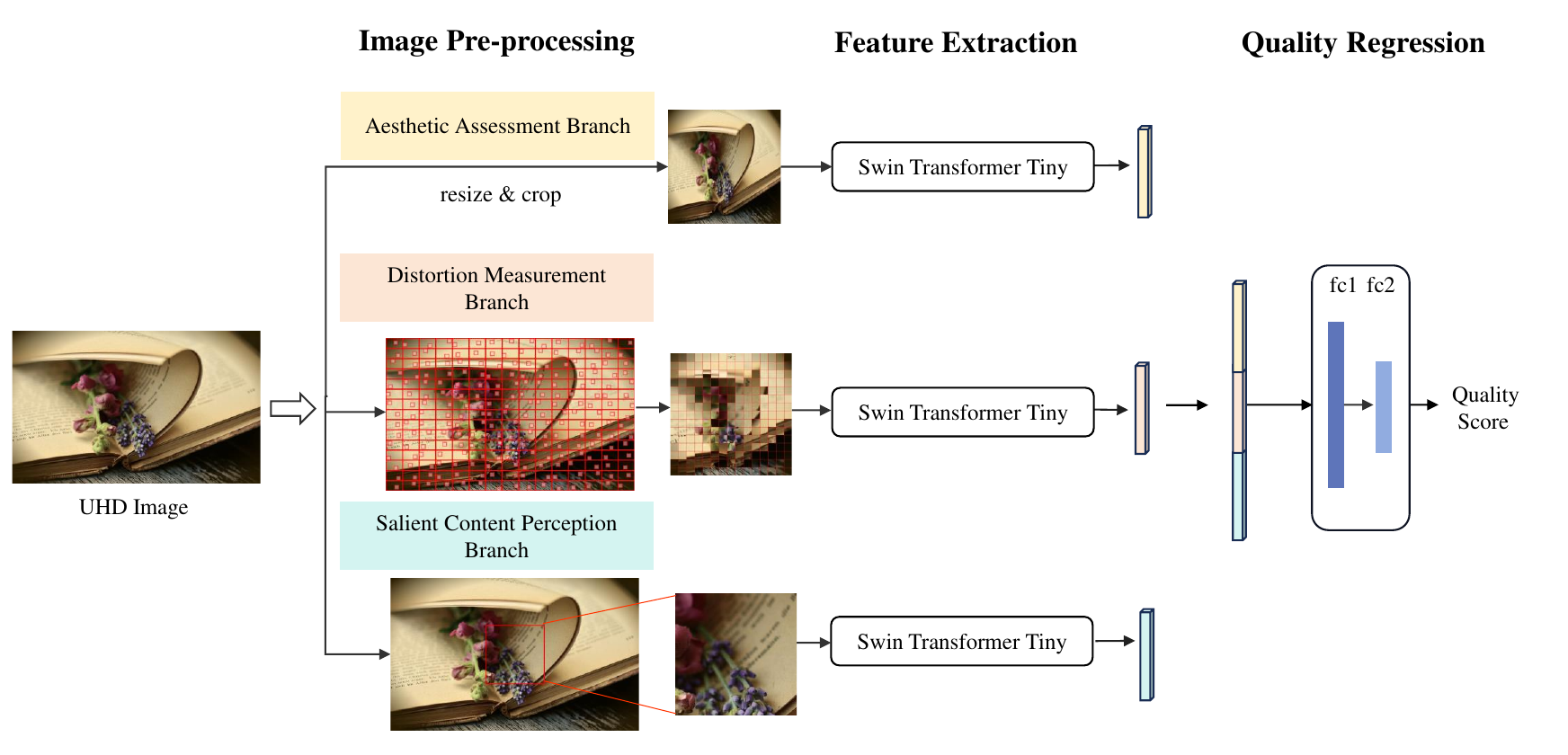}
    \caption{The diagram of the proposed model. It consists of three modules: the image pre-processing module, the feature extraction module, and the quality regression module. We assess the quality of UHD images from three perspectives: global aesthetic characteristics, local technical distortions, and salient content perception, which are evaluated by the aesthetic assessment branch, distortion measurement branch, and salient content perception branch, respectively.}
    \label{fig:model_diagram}
\end{figure*}

As illustrated in Fig.~\ref{fig:model_diagram}, our method is composed of three modules: the image pre-processing module, the feature extraction module, and the quality regression module.

\subsection{Image Pre-processing Module}

The high-resolution characteristics of UHD images introduce significant computational complexity for IQA methods. Therefore, how to downsample high-resolution images into low-resolution ones in the proper manner is crucial for developing efficient UHD IQA methods. The motivation of the proposed method is to decompose the overall quality perception into three dimensions: aesthetic characteristics, technical distortions, and salient content perception. For each dimension, we perform an appropriate downsampling strategy to decrease the resolution while preserving the quality characteristics specific to that dimension.

Specifically, given a UHD image $\bm x\in\mathbb{R}^{H\times W \times 3}$, where $H$ and $W$ represent the height and the width, respectively, with the minimum of $H$ and $W$ being equal to or larger than $2160$. For aesthetic feature extraction, we observe that the factors that affect aesthetic quality are normally global features, such as image content, color, layout, luminance, etc. Therefore, downsampling the UHD image directly into a low-resolution one does not significantly impact the extraction of aesthetic features. We denote this process as:

\begin{equation}
\begin{aligned}
\bm x_{l} &= {\rm D}(\bm x),
\end{aligned}
\end{equation}
where $\rm D(\cdot)$ represents the bilinear downsampling operator and $\bm x_{l}\in\mathbb{R}^{H_l\times W_l \times 3}$ is the low-resolution one of $\bm x$. Note that $H_l \ll H$ and $W_l \ll W$.

In contrast, technical distortions, such as noise and texture, are typically low-level features that are sensitive to the original resolution. So, the downsampling process can cause a substantial loss of detail, thus affecting distortion feature extraction. To overcome this problem, we adopt the grid mini-patch sampling strategy~\cite{wu2022fast} to uniformly sample local mini-patches from the original resolution image. To be more specific, we first uniformly split the UHD image $\bm x$ into $N_g \times N_g$ non-overlapped and equal-size image patches $\bm G = \{{\bm g}_{0,0},...,{\bm g}_{i,j},..., {\bm g}_{N_g-1,N_g - 1}\}$, where ${\bm g}_{i,j}$ denotes the patch in the $i$-th row and $j$-th column. We formulate this process as:
\begin{equation}
\begin{aligned}
\bm g_{i,j} &= \bm x[\frac{i\times H}{N_g}:\frac{(i+1)\times H}{N_g},\frac{j\times W}{N_g}:\frac{(j+1)\times W}{N_g}], \\
\end{aligned}
\end{equation}
where $0 \leq i,j < N_g$.
To preserve the local distortion information, we randomly crop a mini-patch $\bm p_{i,j}$ with the resolution of $H_p\times W_p$ from the patch $\bm g_{i,j}$, where $H_p \ll H/N_g$ and $W_p \ll W/N_g$. Then, we splice these mini-patches into a new fragment image $\bm x_f$ according to their original position:
\begin{equation}
\begin{aligned}
\bm x_f^{i,j} &= \bm x_f[i\times H_p:(i+1)\times H_p,j\times W_p:(j+1)\times W_p], \\
&=\bm p_{i,j}, \quad \quad \quad \quad \quad \quad \quad \quad \quad \quad \quad \quad 0 \leq i,j < N_g. \\
\end{aligned}
\end{equation}
We utilize the fragment image $\bm x_{f}$ with the height of $H_p\times N_g$ and the width of $W_p\times N_g$ for the distortion feature extraction.

To effectively extract features of salient content, it is essential to select the most salient region of the image. Although numerous saliency detection methods have been proposed in the literature to tackle the task, we observe that state-of-the-art saliency detection methods significantly increase computational complexity. Therefore, we choose a simpler approach by leveraging the center bias effect in saliency detection studies~\cite{borji2012state}, which crops the center patch with the resolution of $H_s\times W_s$ from the UHD image:
\begin{equation}
\begin{aligned}
\bm x_s &= \bm x[\frac{H-H_s}{2}:\frac{H+H_s}{2},\frac{W-W_s}{2}:\frac{W+W_s}{2}], \\
\end{aligned}
\end{equation}
where $\bm x_s$ represents the salient image patch of the image $\bm x$ and is used for extracting features of salient content.

\subsection{Feature Extraction Module}
As illustrated in Fig.~\ref{fig:model_diagram}, we design a multi-branch network to extract features of different dimensions. Specifically, we use Swin Transformer Tiny~\cite{liu2021swin} as the backbone network due to its excellent feature extraction capabilities as well as low computational complexity. Although previous studies~\cite{sun2023blind,sun2022deep,sun2019mc360iqa,sun2021deep} suggest that incorporating features extracted from intermediate layers of the backbone network can enhance quality-aware feature representation, we opt for the standard Swin Transformer to keep the model as simple as possible, making it more easily deployable on mobile or edge devices. To further enhance the feature representation, we pre-train the backbone network on the large-scale aesthetic quality assessment dataset AVA~\cite{murray2012ava}. Since each branch of the proposed model is responsible for extracting different types of features, the branches do not share model weights. We formulate the feature extraction process as follows:
\begin{equation}
\begin{aligned}
\mathcal{F}_{\rm aes} &= {\rm Swin}_{\rm aes}(\bm x_l ), \\
\mathcal{F}_{\rm dis} &= {\rm Swin}_{\rm dis}(\bm x_f ), \\
\mathcal{F}_{\rm sal} &= {\rm Swin}_{\rm sal}(\bm x_s ), \\
\end{aligned}
\end{equation}
where $\mathcal{F}_{\rm aes}$, $\mathcal{F}_{\rm dis}$, and $\mathcal{F}_{\rm sal}$ are the extracted aesthetic, technical distortion, and salient content features by the feature extractors ${\rm Swin}_{\rm aes}$, ${\rm Swin}_{\rm dis}$, and ${\rm Swin}_{\rm sal}$, respectively.

\subsection{Quality Regression Module}
After extracting three dimension features, we concatenate them into the final feature representation $\mathcal{F}$: 
\begin{equation}
\begin{aligned}
&\mathcal{F} = {\rm CAT}(\mathcal{F}_{\rm aes}, \mathcal{F}_{\rm dis}, \mathcal{F}_{\rm sal}), \\
\end{aligned}
\end{equation}
where $\rm CAT$ denotes the concatenation operator.

We then utilize a MLP network to regress $\mathcal{F}$ into the overall quality scores $\hat{q}$:
\begin{equation}
\begin{aligned}
 \hat{q} &= {\rm MLP}(\mathcal{F}), \\
\end{aligned}
\end{equation}
where $\rm MLP$ denotes the two-layer MLP network consisting of $128$ and $1$ neurons.

\subsection{Loss Function}
Mean square error (MSE) and mean absolute error (MAE) are two common loss functions used to minimize the discrepancy between model quality scores $\hat{q}$ and ground-truth quality scores $q$. However, MSE and MAE cannot optimize the rank consistency between image pairs, which is crucial in real-world applications. Therefore, we adopt a combination of MSE loss and fidelity loss to optimize the proposed model, where fidelity loss~\cite{tsai2007frank} targets the rank consistency between image pairs.

Specifically, we randomly sample two images ($\bm{x}$, $\bm{y}$) as a pair and calculate the quality scores of two images $\hat{q}(\bm{x})$ and $\hat{q}(\bm{y})$ using the proposed model. We then calculate a binary label based on their ground-truth mean opinion scores (MOSs):
\begin{equation}
    p(\bm{x}, \bm{y}) = \left\{ 
    \begin{aligned} 
    & 0 \quad {\rm{if}} \quad q(\bm{x}) \geq q(\bm{y})\\
    & 1 \quad \rm{otherwise}
    \end{aligned}
    \right.
    ,
\end{equation}
where $q(\bm{x})$ and $q(\bm{y})$ are the ground-truth MOSs of the image pair ($\bm{x}$, $\bm{y}$).
We estimate the probability of $\bm{x}$ perceived better than $\bm{y}$ as

\begin{equation}
\hat{p}(\bm{x}, \bm{y}) = \Phi(\frac{\hat{q}(\bm{x})-\hat{q}(\bm{y})}{\sqrt{2}}),
\end{equation}
where $\Phi(\cdot)$ is the standard Normal cumulative distribution function, and the variance is fixed to one. We then calculate fidelity loss for this pair:
\begin{equation}
\begin{aligned}
\ell_{\rm fidelity}(\bm{x}, \bm{y};\bm{\theta}) = &1-\sqrt{p(\bm{x}, \bm{y})\hat{p}(\bm{x}, \bm{y})}\\
&-\sqrt{(1-p(\bm{x}, \bm{y}))(1-\hat{p}(\bm{x}, \bm{y}))},
\end{aligned}
\end{equation}
where $\theta$ represents the parameters of the proposed model. MSE loss are calculated as:
\begin{equation}
   \ell_{\rm MSE}(\bm{x}, \bm{y};\bm{\theta}) = (q(\bm x) - \hat{q}(\bm x))^2 + (q(\bm y) - \hat{q}(\bm y))^2.
\end{equation}

The overall loss is calculated as the weighted sum of fidelity loss and MSE loss:
\begin{equation}
   \ell(\bm{x}, \bm{y};\bm{\theta}) = \alpha \ell_{\rm fidelity}(\bm{x}, \bm{y};\bm{\theta}) + 
\beta \ell_{\rm MSE}(\bm{x}, \bm{y};\bm{\theta}),
\end{equation}
where $\alpha$ and $\beta$ are weights of fidelity loss and MSE loss respectively.

\begin{table}[tb]
  \caption{Performance of the compared models and the proposed model on the validation set of the UHD-IQA dataset. The best-performing model is highlighted in each column
  }
  \label{tab:validation}
  \centering
  \begin{tabular}{lcccccc}
    \toprule
    Methods & SRCC $\uparrow$ & PLCC $\uparrow$ & KRCC $\uparrow$ & RMSE $\downarrow$ & MAE $\downarrow$ & MACs (G) $\downarrow$ \\
    \midrule
    HyperIQA~\cite{su2020blindly}  & $0.524$ &  $0.182$ & $0.359$& $0.087$& $0.055$&$211$\\
    Effnet-2C-MLSP~\cite{wiedemann2023konx} & $0.615$ & $0.627$ &$0.445$ & $0.060$& $0.050$&$345$\\
    CONTRIQUE~\cite{madhusudana2022image} & $0.716$ & $0.712$& $0.521$& $0.049$& $0.038$&  $855$\\
    ARNIQA~\cite{agnolucci2024arniqa} & $0.718$ &  $0.717$ &  $0.523$& $0.050$& $0.039$&$855$\\
    CLIP-IQA+~\cite{wang2023exploring} & $0.743$ & $0.732$& $0.546$& $0.108$& $0.087$&$895$ \\
    QualiCLIP~\cite{agnolucci2024quality} & $0.757$ & $0.752$& $0.557$& $0.079$& $0.064$&$901$ \\
    \cdashline{1-7}
    Proposed & $\textbf{0.817}$ & $\textbf{0.823}$& $\textbf{0.625}$& $\textbf{0.040}$& $\textbf{0.032}$&$\textbf{43.5}$ \\
  \bottomrule
  \end{tabular}
\end{table}


\section{Experiment}
\subsection{Experimental Protocol}

\noindent\textbf{Implementation Details.} We utilize Swin Transformer Tiny~\cite{liu2021swin} as the backbone network of the three branches. To enhance its quality-aware feature representation capability, we first pre-train it on the AVA dataset~\cite{murray2012ava}, following the same training strategy as the proposed model. For the aesthetic branch, we resize the resolution of the minimum dimension of the UHD image as $512$ pixels (\textit{i.e.,} $\min(H_l,W_l)= 512$) while preserving its aspect ratios. During the training and test stages, we randomly crop and centrally crop the resized image to a resolution of $480\times480$ as the input, respectively. For the distortion branch, the number of mini-patches $N_g\times N_g$ is set as $15\times 15$, with each mini-patch having a resolution of $32\times 32$ (\textit{i.e.,}$H_p =  W_p = 32$). Thus, the resolution of the fragment image is also $480\times 480$. For the salient perception branch, the resolution of the center patch is set to $480\times 480$. The Adam optimizer, with an initial learning rate $1\times10^{-5}$ and a batch size $12$, is used to train the proposed model on a server with $2$ NVIDIA RTX 3090 GPUs. The learning rate is decayed by a factor of $10$ after $10$ epochs, with a total of $100$ epochs. The weights of fidelity loss $\alpha$ and MSE loss $\beta$ are set to $1$ and $0.1$ respectively.

\begin{table}[tb]
  \caption{Performance of the compared models and the proposed model on the test set of the UHD-IQA dataset. The best-performing model is highlighted in each column
  }
  \label{tab:test}
  \centering
  \begin{tabular}{lcccccc}
    \toprule
    Methods & SRCC $\uparrow$ & PLCC $\uparrow$ & KRCC $\uparrow$ & RMSE $\downarrow$ & MAE $\downarrow$ & MACs (G) $\downarrow$ \\
    \midrule
    HyperIQA~\cite{su2020blindly}  & $ 0.553$ &  $ 0.103$ & $ 0.389$& $0.118$& $ 0.070$&$211$\\
    Effnet-2C-MLSP~\cite{wiedemann2023konx} & $0.675$ & $0.641$ &$ 0.491$ & $0.074$& $0.059$&$345$\\
    CONTRIQUE~\cite{madhusudana2022image} & $0.732$ & $0.678$& $ 0.532$& $0.073$& $0.052$&  $855$\\
    ARNIQA~\cite{agnolucci2024arniqa} & $0.739$ &  $0.694$ &  $0.544$& $0.052$& $0.739$&$855$\\
    CLIP-IQA+~\cite{wang2023exploring} & $0.747$ & $0.709$& $0.551$& $0.111$& $0.089$&$895$ \\
    QualiCLIP~\cite{agnolucci2024quality} & $0.770$ & $0.725$& $0.570$& $0.083$& $0.066$&$901$ \\
    \cdashline{1-7}
    Proposed & $\textbf{0.846}$ & $\textbf{0.798}$& $\textbf{0.657}$& $\textbf{0.061}$& $\textbf{0.042}$&$\textbf{43.5}$ \\
  \bottomrule
  \end{tabular}
\end{table}

\noindent\textbf{Test Datasets.} We evaluate the proposed model on the largest UHD IQA dataset, UHD-IQA~\cite{hosu2024uhd}, which focuses on highly aesthetic photos of high technical quality. UHD-IQA comprises $6,073$ images, with $70\%$, $15\%$, and $15\%$ for the training, validation, and test sets, respectively. Note that only the labels of the training set are accessible in the \href{https://codalab.lisn.upsaclay.fr/competitions/19335}{AIM 2024 UHD-IQA Challenge: Pushing the Boundaries of Blind Photo Quality Assessment}~\footnote{https://codalab.lisn.upsaclay.fr/competitions/19335}. Therefore, we randomly split the public training set into $80\%$ and $20\%$ for the training-private and validation-private sets respectively, which are used to train the proposed model and select the best one. We report the performance on the validation and test sets for model comparison.

\noindent\textbf{Compared Models.} We compare the proposed method with six popular IQA methods, including HyperIQA~\cite{su2020blindly}, Effnet-2C-MLSP~\cite{wiedemann2023konx}, CONTRIQUE~\cite{madhusudana2022image}, ARNIQA~\cite{agnolucci2024arniqa}, CLIP-IQA+~\cite{wang2023exploring}, and QualiCLIP~\cite{agnolucci2024quality}.

\noindent\textbf{Evaluation Criteria.}
We use five criteria to evaluate the performance of IQA models: Spearman Rank-order Correlation Coefficient (SRCC), Pearson’s Linear Correlation Coefficient (PLCC), Kendall’s Rank Correlation Coefficient (KRCC), Root Mean Squared Error (RMSE), and Mean Absolute Error (MAE). 

\subsection{Experimental Results}

We present the performance of IQA models on the validation set in Table~\ref{tab:validation} and on the test set in Table~\ref{tab:test}, from which several conclusions can be drawn. First, we observe that existing IQA methods have high computational complexity for UHD images, with MACs exceeding $200$G. In contrast, our method has significantly lower computational complexity, with $43.5$G MACs~\footnote{It takes 0.068 seconds to calculate the quality score of one image on an NVIDIA RTX 3090 GPU.}, which is nearly four times less than the fastest IQA method. Second, despite high computational complexity, existing IQA methods still perform poorly on the UHD-IQA dataset, indicating that these methods are unable to effectively capture high-resolution image quality representation. Third, the proposed method achieves the best performance on both the validation and test sets, outperforming the second-best method, QualiCLIP, by $8\%$ on the validation set and $10\%$ on the test set in terms of the SRCC metric, while reducing computational complexity by $20$ times, which demonstrates its effectiveness in assessing the quality of UHD images.

\begin{table}[tb]
  \caption{The results of AIM 2024 UHD-IQA Challenge
  }
  \label{tab:challenge}
  \centering
  \begin{tabular}{lccccc}
    \toprule
    Methods & SRCC $\uparrow$ & PLCC $\uparrow$ & KRCC $\uparrow$ & RMSE $\downarrow$ & MAE $\downarrow$  \\
    \midrule
    SJTU MMLab (ours) & $\textbf{0.846}$ & $0.798$& $\textbf{0.657}$& $\textbf{0.061}$& $\textbf{0.042}$ \\
    CIPLAB & $0.835$ & $\textbf{0.800}$ &$ 0.642$ & $0.064$& $0.044$\\
    ZX AIE Vector & $0.795$ & $0.768$& $ 0.605$& $0.062$& $0.044$\\
    $I^2$Group & $0.788$ &  $0.756$ &  $0.598$& $0.066$& $0.046$\\
    Dominator & $0.731$ & $0.712$& $0.539$& $0.072$& $0.052$ \\
    ICL & $0.517$ & $0.521$& $0.361$& $0.136$& $0.115$ \\
  \bottomrule
  \end{tabular}
\end{table}

We list the results of AIM 2024 UHD-IQA Challenge in Table~\ref{tab:challenge}. From Table~\ref{tab:challenge}, it is shown that the proposed model significantly outperforms other competing teams, which further demonstrates the superiority of the proposed method.

\begin{table}[tb]
  \caption{The results of ablation studies on model design
  }
  \label{tab:branch}
  \centering
  \begin{tabular}{ccc|ccccc}
    \toprule
    Aesthetics & Distortions & Saliency & SRCC $\uparrow$ & PLCC $\uparrow$ & KRCC $\uparrow$ & RMSE $\downarrow$ & MAE $\downarrow$  \\
    \midrule
$\surd$&$\times$ &$\times$ &$0.721$ & $0.738$& $0.535$& $0.047$& $0.037$ \\
$\times$& $\surd$&$\times$ & $0.791$&$0.794$ &$0.598$ & $0.043$&$0.034$  \\
$\times$& $\times$& $\surd$& $0.696$& $0.702$& $0.506$&$0.050$ & $0.039$ \\
$\surd$& $\surd$&$\times$ & $0.811$& $0.820$& $0.619$& $0.041$& $0.032$ \\
$\surd$& $\times$& $\surd$& $0.782$& $0.793$& $0.588$& $0.043$& $0.034$ \\
$\times$& $\surd$& $\surd$& $0.787$& $0.792$& $0.594$& $0.044$& $0.034$ \\
    $\surd$& $\surd$& $\surd$& $\textbf{0.817}$ & $\textbf{0.823}$& $\textbf{0.625}$& $\textbf{0.040}$& $\textbf{0.032}$ \\
  \bottomrule
  \end{tabular}
\end{table}

\begin{table}[tb]
  \caption{The results of ablation studies on backbone networks and pre-trained datasets
  }
  \label{tab:backbone}
  \centering
  \begin{tabular}{ll|ccccc}
    \toprule
    Backbones & Pre-trained  & SRCC $\uparrow$ & PLCC $\uparrow$ & KRCC $\uparrow$ & RMSE $\downarrow$ & MAE $\downarrow$  \\
    \midrule
MobileNet v2 & ImageNet-1k & $0.625$ & $0.631$& $0.444$& $0.054$&$0.043$  \\
ResNet-50&ImageNet-1k &  $0.600$& $0.613$& $0.427$& $0.054$& $0.043$ \\
Swin-T& ImageNet-1k&  $0.789$& $0.791$& $0.593$& $0.044$& $0.034$ \\
Swin-T& AVA &  $0.817$& $0.823$& $0.625$& $0.040$&  $0.032$\\
Swin-B& AVA & $\textbf{0.830}$ & $\textbf{0.836}$& $\textbf{0.639}$& $\textbf{0.039}$& $\textbf{0.031}$ \\
  \bottomrule
  \end{tabular}
\end{table}

\subsection{Ablation Studies}
\noindent\textbf{Network Design.} The proposed model consists of three branches for extracting aesthetic, distortion, and salient content features. To validate the effectiveness of these branches, we train each individual branch and combinations of two branches on the UHD-IQA dataset. The performance on the validation set is listed in Table~\ref{tab:branch}. For the single-branch network comparison, the distortion branch achieves the best performance, followed by the aesthetic branch, while the salient content perception branch performs moderately. This indicates that the distortion branch plays the most critical role in evaluating the quality of UHD images, suggesting that the grid mini-patch sampling strategy is effective for addressing the UHD IQA task. For the two-branch network comparison, the combination of the aesthetic branch and the distortion branch performs the best, followed by the combination of the distortion branch and the salient branch, and the combination of the aesthetic branch and the salient branch is slightly inferior to the combination of the distortion branch and the salient branch. This further confirms that the distortion branch contributes the most to the proposed model. 

\noindent\textbf{Backbones and pre-trained dataset.} We investigate the effectiveness of different backbone networks and pre-trained datasets. Specifically, we evaluate two CNN networks: MobileNet v2~\cite{sandler2018mobilenetv2} and ResNet-50~\cite{he2016deep} pre-trained on ImageNet-1k~\cite{deng2009imagenet}. For the default backbone, Swin Transformer Tiny, we also evaluate its performance when pre-trained on ImageNet-1k and its larger counterpart, Swin Transformer Base, pre-trained on AVA. The results are presented in Table~\ref{tab:backbone}. First, by comparing the performance of Swin Transformer Tiny pre-trained on ImageNet-1k and AVA, we observe that per-trained on AVA boosts model performance, indicating that AVA pre-training can significantly enhance the quality-aware feature representation ability. Second, the performance of MobileNet v2 and ResNet-50 is significantly inferior to that of Swin Transformer Tiny, which may be because CNN is difficult to handle fragment images due to the discontinuity characteristics of fragment images. Third, scaling up the Swin Transformer Tiny into the Base leads to a performance improvement, indicating a more powerful backbone network helps improve evaluation ability.

\noindent\textbf{Resolutions.} We study the impact of the resolution of inputs on model performance by setting the inputs of three branches to $224\times224$ and $384\times384$. We train the proposed model with different input resolutions on the UHD-IQA datasets and list the results in Table~\ref{tab:resolution}. We observe that increasing the input resolutions consistently improves model performance, because higher resolutions contain more information for feature extraction. Despite this, the largest resolution in Table~\ref{tab:resolution} is still far less than the original resolution of UHD images, demonstrating the model's efficiency in assessing the quality of UHD images.

\noindent\textbf{Loss functions.} The proposed model is optimized by a combination of fidelity loss and MSE loss. The effectiveness of each individual loss function is evaluated in Table~\ref{tab:loss_function}. The results show that fidelity loss and MSE loss individually perform similarly but are both inferior to their combination, highlighting the effectiveness of the combined loss function.

\begin{table}[tb]
  \caption{The results of ablation studies on input resolutions
  }
  \label{tab:resolution}
  \centering
  \begin{tabular}{l|cccccc}
    \toprule
    Resolution   & SRCC $\uparrow$ & PLCC $\uparrow$ & KRCC $\uparrow$ & RMSE $\downarrow$ & MAE $\downarrow$  \\
    \midrule
$224\times224$  & $0.765$& $0.773$& $0.570$& $0.045$& $0.035$ \\
$384\times384$ & $0.809$ & $0.810$& $0.614$& $0.042$& $0.033$ \\
$480\times480$&  $\textbf{0.817}$ & $\textbf{0.823}$& $\textbf{0.625}$& $\textbf{0.040}$& $\textbf{0.032}$  \\
  \bottomrule
  \end{tabular}
\end{table}

\begin{table}[tb]
  \caption{The results of ablation studies on loss function
  }
  \label{tab:loss_function}
  \centering
  \begin{tabular}{l|cccccc}
    \toprule
    Loss Function   & SRCC $\uparrow$ & PLCC $\uparrow$ & KRCC $\uparrow$ & RMSE $\downarrow$ & MAE $\downarrow$  \\
    \midrule
Fidelity loss  & $0.804$& $0.814$& $0.613$& $0.042$& $0.033$ \\
MSE loss & $0.804$ & $0.813$& $0.614$& $0.042$& $0.033$ \\
Combination &  $\textbf{0.817}$ & $\textbf{0.823}$& $\textbf{0.625}$& $\textbf{0.040}$& $\textbf{0.032}$  \\
  \bottomrule
  \end{tabular}
\end{table}

\section{Conclusion}
In this paper, we design a multi-branch DNN model to evaluate the quality of UHD images from three perspectives: global aesthetic characteristics, local technical distortions, and salient content perception, while avoiding direct processing of high-resolution images. By decomposing the overall quality assessment of a high-resolution image into three quality dimension measurements of the low-resolution ones, our method effectively assesses the quality of UHD images with acceptable computational complexity. What's more, we avoid complex model designs and use only the standard DNN structures, making it easy to implement in practical applications and optimize for hardware.

\section*{Acknowledgements}
This work was supported in part by the National Natural Science Foundation of China under Grants 62301316, 62371283, 62225112, and 62271312, the China Postdoctoral Science Foundation under Grants 2023TQ0212 and 2023M742298, the Postdoctoral Fellowship Program of CPSF under Grant GZC20231618, the Fundamental Research Funds for the Central Universities, the National Key R\&D Program of China (2021YFE0206700), the Science and Technology Commission of Shanghai Municipality (2021SHZDZX0102), and the Shanghai Committee of Science and Technology (22DZ2229005).

%
%
\bibliographystyle{splncs04}
\bibliography{main}

\begin{thebibliography}{10}
\providecommand{\url}[1]{\texttt{#1}}
\providecommand{\urlprefix}{URL }
\providecommand{\doi}[1]{https://doi.org/#1}

\bibitem{agnolucci2024quality}
Agnolucci, L., Galteri, L., Bertini, M.: Quality-aware image-text alignment for real-world image quality assessment. arXiv preprint arXiv:2403.11176  (2024)

\bibitem{agnolucci2024arniqa}
Agnolucci, L., Galteri, L., Bertini, M., Del~Bimbo, A.: Arniqa: Learning distortion manifold for image quality assessment. In: Proceedings of the IEEE/CVF Winter Conference on Applications of Computer Vision. pp. 189--198 (2024)

\bibitem{borji2012state}
Borji, A., Itti, L.: State-of-the-art in visual attention modeling. IEEE Transactions on Pattern Analysis and Machine Intelligence  \textbf{35}(1),  185--207 (2012)

\bibitem{bosse2017deep}
Bosse, S., Maniry, D., M{\"u}ller, K.R., Wiegand, T., Samek, W.: Deep neural networks for no-reference and full-reference image quality assessment. IEEE Transactions on Image Processing  \textbf{27}(1),  206--219 (2017)

\bibitem{chen2024topiq}
Chen, C., Mo, J., Hou, J., Wu, H., Liao, L., Sun, W., Yan, Q., Lin, W.: Topiq: A top-down approach from semantics to distortions for image quality assessment. IEEE Transactions on Image Processing  (2024)

\bibitem{deng2009imagenet}
Deng, J., Dong, W., Socher, R., Li, L.J., Li, K., Fei-Fei, L.: Imagenet: A large-scale hierarchical image database. In: 2009 IEEE conference on computer vision and pattern recognition. pp. 248--255. Ieee (2009)

\bibitem{gu2014using}
Gu, K., Zhai, G., Yang, X., Zhang, W.: Using free energy principle for blind image quality assessment. IEEE Transactions on Multimedia  \textbf{17}(1),  50--63 (2014)

\bibitem{he2016deep}
{He}, K., {Zhang}, X., {Ren}, S., {Sun}, J.: Deep residual learning for image recognition. In: 2016 IEEE Conference on Computer Vision and Pattern Recognition (CVPR). pp. 770--778 (2016)

\bibitem{hosu2024uhd}
Hosu, V., Agnolucci, L., Wiedemann, O., Iso, D.: Uhd-iqa benchmark database: Pushing the boundaries of blind photo quality assessment. arXiv preprint arXiv:2406.17472  (2024)

\bibitem{hosu2020koniq}
Hosu, V., Lin, H., Sziranyi, T., Saupe, D.: Koniq-10k: An ecologically valid database for deep learning of blind image quality assessment. IEEE Transactions on Image Processing  \textbf{29},  4041--4056 (2020)

\bibitem{huang2024high}
Huang, H., Wan, Q., Korhonen, J.: High resolution image quality database. In: ICASSP 2024-2024 IEEE International Conference on Acoustics, Speech and Signal Processing (ICASSP). pp. 3105--3109. IEEE (2024)

\bibitem{kang2014convolutional}
Kang, L., Ye, P., Li, Y., Doermann, D.: Convolutional neural networks for no-reference image quality assessment. In: Proceedings of the IEEE conference on computer vision and pattern recognition. pp. 1733--1740 (2014)

\bibitem{kao2017deep}
Kao, Y., He, R., Huang, K.: Deep aesthetic quality assessment with semantic information. IEEE Transactions on Image Processing  \textbf{26}(3),  1482--1495 (2017)

\bibitem{ke2021musiq}
Ke, J., Wang, Q., Wang, Y., Milanfar, P., Yang, F.: Musiq: Multi-scale image quality transformer. In: Proceedings of the IEEE/CVF International Conference on Computer Vision. pp. 5148--5157 (2021)

\bibitem{korhonen2021consumer}
Korhonen, J., Su, Y., You, J.: Consumer image quality prediction using recurrent neural networks for spatial pooling. arXiv preprint arXiv:2106.00918  (2021)

\bibitem{li2022full}
Li, C., Zhang, Z., Sun, W., Min, X., Zhai, G.: A full-reference quality assessment metric for cartoon images. In: 2022 IEEE 24th International Workshop on Multimedia Signal Processing (MMSP). pp.~1--6. IEEE (2022)

\bibitem{liu2019unsupervised}
Liu, Y., Gu, K., Zhang, Y., Li, X., Zhai, G., Zhao, D., Gao, W.: Unsupervised blind image quality evaluation via statistical measurements of structure, naturalness, and perception. IEEE Transactions on Circuits and Systems for Video Technology  \textbf{30}(4),  929--943 (2019)

\bibitem{liu2021swin}
Liu, Z., Lin, Y., Cao, Y., Hu, H., Wei, Y., Zhang, Z., Lin, S., Guo, B.: Swin transformer: Hierarchical vision transformer using shifted windows. In: Proceedings of the IEEE/CVF international conference on computer vision. pp. 10012--10022 (2021)

\bibitem{lu2022deep}
Lu, W., Sun, W., Min, X., Zhu, W., Zhou, Q., He, J., Wang, Q., Zhang, Z., Wang, T., Zhai, G.: Deep neural network for blind visual quality assessment of 4k content. IEEE Transactions on Broadcasting  \textbf{69}(2),  406--421 (2022)

\bibitem{ma2017end}
Ma, K., Liu, W., Zhang, K., Duanmu, Z., Wang, Z., Zuo, W.: End-to-end blind image quality assessment using deep neural networks. IEEE Transactions on Image Processing  \textbf{27}(3),  1202--1213 (2017)

\bibitem{madhusudana2022image}
Madhusudana, P.C., Birkbeck, N., Wang, Y., Adsumilli, B., Bovik, A.C.: Image quality assessment using contrastive learning. IEEE Transactions on Image Processing  \textbf{31},  4149--4161 (2022)

\bibitem{min2016blind}
Min, X., Zhai, G., Gu, K., Fang, Y., Yang, X., Wu, X., Zhou, J., Liu, X.: Blind quality assessment of compressed images via pseudo structural similarity. In: 2016 IEEE International Conference on Multimedia and Expo (ICME). pp.~1--6. IEEE (2016)

\bibitem{min2018blind}
{Min}, X., {Zhai}, G., {Gu}, K., {Liu}, Y., {Yang}, X.: Blind image quality estimation via distortion aggravation. IEEE Transactions on Broadcasting  \textbf{64}(2),  508--517 (2018)

\bibitem{mittal2012no}
Mittal, A., Moorthy, A.K., Bovik, A.C.: No-reference image quality assessment in the spatial domain. IEEE Transactions on image processing  \textbf{21}(12),  4695--4708 (2012)

\bibitem{mittal2012making}
Mittal, A., Soundararajan, R., Bovik, A.C.: Making a “completely blind” image quality analyzer. IEEE Signal processing letters  \textbf{20}(3),  209--212 (2012)

\bibitem{moorthy2011blind}
Moorthy, A.K., Bovik, A.C.: Blind image quality assessment: From natural scene statistics to perceptual quality. IEEE transactions on Image Processing  \textbf{20}(12),  3350--3364 (2011)

\bibitem{murray2012ava}
{Murray}, N., {Marchesotti}, L., {Perronnin}, F.: Ava: A large-scale database for aesthetic visual analysis. In: 2012 IEEE Conference on Computer Vision and Pattern Recognition. pp. 2408--2415 (2012)

\bibitem{saad2012blind}
Saad, M.A., Bovik, A.C., Charrier, C.: Blind image quality assessment: A natural scene statistics approach in the dct domain. IEEE transactions on Image Processing  \textbf{21}(8),  3339--3352 (2012)

\bibitem{saha2023re}
Saha, A., Mishra, S., Bovik, A.C.: Re-iqa: Unsupervised learning for image quality assessment in the wild. In: Proceedings of the IEEE/CVF conference on computer vision and pattern recognition. pp. 5846--5855 (2023)

\bibitem{sandler2018mobilenetv2}
Sandler, M., Howard, A., Zhu, M., Zhmoginov, A., Chen, L.C.: Mobilenetv2: Inverted residuals and linear bottlenecks. In: Proceedings of the IEEE conference on computer vision and pattern recognition. pp. 4510--4520 (2018)

\bibitem{shah2021real}
Shah, R.R., Akundy, V.A., Wang, Z.: Real versus fake 4k-authentic resolution assessment. In: ICASSP 2021-2021 IEEE International Conference on Acoustics, Speech and Signal Processing (ICASSP). pp. 2185--2189. IEEE (2021)

\bibitem{su2020blindly}
Su, S., Yan, Q., Zhu, Y., Zhang, C., Ge, X., Sun, J., Zhang, Y.: Blindly assess image quality in the wild guided by a self-adaptive hyper network. In: Proceedings of the IEEE/CVF Conference on Computer Vision and Pattern Recognition. pp. 3667--3676 (2020)

\bibitem{sun2022deep}
Sun, W., Min, X., Lu, W., Zhai, G.: A deep learning based no-reference quality assessment model for ugc videos. In: Proceedings of the 30th ACM International Conference on Multimedia. pp. 856--865 (2022)

\bibitem{sun2023blind}
Sun, W., Min, X., Tu, D., Ma, S., Zhai, G.: Blind quality assessment for in-the-wild images via hierarchical feature fusion and iterative mixed database training. IEEE Journal of Selected Topics in Signal Processing  \textbf{17}(6),  1178--1192 (2023)

\bibitem{sun2019mc360iqa}
Sun, W., Min, X., Zhai, G., Gu, K., Duan, H., Ma, S.: Mc360iqa: A multi-channel cnn for blind 360-degree image quality assessment. IEEE Journal of Selected Topics in Signal Processing  \textbf{14}(1),  64--77 (2019)

\bibitem{sun2021deep}
Sun, W., Wang, T., Min, X., Yi, F., Zhai, G.: Deep learning based full-reference and no-reference quality assessment models for compressed ugc videos. In: 2021 IEEE International Conference on Multimedia \& Expo Workshops (ICMEW). pp.~1--6. IEEE (2021)

\bibitem{sun2024analysis}
Sun, W., Wen, W., Min, X., Lan, L., Zhai, G., Ma, K.: Analysis of video quality datasets via design of minimalistic video quality models. IEEE Transactions on Pattern Analysis and Machine Intelligence  (2024)

\bibitem{sun2024enhancing}
Sun, W., Wu, H., Zhang, Z., Jia, J., Zhang, Z., Cao, L., Chen, Q., Min, X., Lin, W., Zhai, G.: Enhancing blind video quality assessment with rich quality-aware features. arXiv preprint arXiv:2405.08745  (2024)

\bibitem{sun2024dual}
Sun, W., Zhang, W., Jiang, Y., Wu, H., Zhang, Z., Jia, J., Zhou, Y., Ji, Z., Min, X., Lin, W., et~al.: Dual-branch network for portrait image quality assessment. arXiv preprint arXiv:2405.08555  (2024)

\bibitem{tan2024highly}
Tan, X., Zhang, J., Quan, Y., Li, J., Wu, Y., Bian, Z.: Highly efficient no-reference 4k video quality assessment with full-pixel covering sampling and training strategy. arXiv preprint arXiv:2407.20766  (2024)

\bibitem{tsai2007frank}
Tsai, M.F., Liu, T.Y., Qin, T., Chen, H.H., Ma, W.Y.: Frank: a ranking method with fidelity loss. In: Proceedings of the 30th annual international ACM SIGIR conference on Research and development in information retrieval. pp. 383--390 (2007)

\bibitem{wang2023exploring}
Wang, J., Chan, K.C., Loy, C.C.: Exploring clip for assessing the look and feel of images. In: Proceedings of the AAAI Conference on Artificial Intelligence. vol.~37, pp. 2555--2563 (2023)

\bibitem{wang2022deep}
Wang, T., Sun, W., Wu, W., Chen, Y., Min, X., Lu, W., Zhang, Z., Zhai, G.: A deep learning-based multidimensional aesthetic quality assessment method for mobile game images. IEEE Transactions on Games  \textbf{15}(4),  658--668 (2022)

\bibitem{wiedemann2023konx}
Wiedemann, O., Hosu, V., Su, S., Saupe, D.: Konx: cross-resolution image quality assessment. Quality and User Experience  \textbf{8}(1), ~8 (2023)

\bibitem{wu2022fast}
Wu, H., Chen, C., Hou, J., Liao, L., Wang, A., Sun, W., Yan, Q., Lin, W.: Fast-vqa: Efficient end-to-end video quality assessment with fragment sampling. In: European Conference on Computer Vision. pp. 538--554 (2022)

\bibitem{wu2023q}
Wu, H., Zhang, Z., Zhang, W., Chen, C., Liao, L., Li, C., Gao, Y., Wang, A., Zhang, E., Sun, W., et~al.: Q-align: Teaching lmms for visual scoring via discrete text-defined levels. arXiv preprint arXiv:2312.17090  (2023)

\bibitem{yang2019sgdnet}
Yang, S., Jiang, Q., Lin, W., Wang, Y.: Sgdnet: An end-to-end saliency-guided deep neural network for no-reference image quality assessment. In: Proceedings of the 27th ACM international conference on multimedia. pp. 1383--1391 (2019)

\bibitem{zhai2020perceptual}
Zhai, G., Min, X.: Perceptual image quality assessment: a survey. Science China Information Sciences  \textbf{63},  1--52 (2020)

\bibitem{zhai2019free}
Zhai, G., Min, X., Liu, N.: Free-energy principle inspired visual quality assessment: An overview. Digital Signal Processing  \textbf{91},  11--20 (2019)

\bibitem{zhai2021perceptual}
Zhai, G., Sun, W., Min, X., Zhou, J.: Perceptual quality assessment of low-light image enhancement. ACM Transactions on Multimedia Computing, Communications, and Applications (TOMM)  \textbf{17}(4),  1--24 (2021)

\bibitem{zhai2011psychovisual}
Zhai, G., Wu, X., Yang, X., Lin, W., Zhang, W.: A psychovisual quality metric in free-energy principle. IEEE Transactions on Image Processing  \textbf{21}(1),  41--52 (2011)

\bibitem{zhang2014vsi}
Zhang, L., Shen, Y., Li, H.: Vsi: A visual saliency-induced index for perceptual image quality assessment. IEEE Transactions on Image processing  \textbf{23}(10),  4270--4281 (2014)

\bibitem{zhang2020blind}
{Zhang}, W., {Ma}, K., {Yan}, J., {Deng}, D., {Wang}, Z.: Blind image quality assessment using a deep bilinear convolutional neural network. IEEE Transactions on Circuits and Systems for Video Technology  \textbf{30}(1),  36--47 (2020)

\bibitem{zhang2021uncertainty}
Zhang, W., Ma, K., Zhai, G., Yang, X.: Uncertainty-aware blind image quality assessment in the laboratory and wild. IEEE Transactions on Image Processing  \textbf{30},  3474--3486 (2021)

\bibitem{zhang2023blind}
Zhang, W., Zhai, G., Wei, Y., Yang, X., Ma, K.: Blind image quality assessment via vision-language correspondence: A multitask learning perspective. In: Proceedings of the IEEE/CVF conference on computer vision and pattern recognition. pp. 14071--14081 (2023)

\bibitem{zhang2021full}
Zhang, Z., Sun, W., Min, X., Wang, T., Lu, W., Zhai, G.: A full-reference quality assessment metric for fine-grained compressed images. In: 2021 International Conference on Visual Communications and Image Processing (VCIP). pp.~1--4. IEEE (2021)

\bibitem{zhang2021no}
Zhang, Z., Sun, W., Min, X., Zhu, W., Wang, T., Lu, W., Zhai, G.: A no-reference evaluation metric for low-light image enhancement. In: 2021 IEEE International Conference on Multimedia and Expo (ICME). pp.~1--6. IEEE (2021)

\bibitem{zhang2022no}
Zhang, Z., Sun, W., Min, X., Zhu, W., Wang, T., Lu, W., Zhai, G.: A no-reference deep learning quality assessment method for super-resolution images based on frequency maps. In: 2022 IEEE International Symposium on Circuits and Systems (ISCAS). pp. 3170--3174. IEEE (2022)

\bibitem{zhao2023quality}
Zhao, K., Yuan, K., Sun, M., Li, M., Wen, X.: Quality-aware pre-trained models for blind image quality assessment. In: Proceedings of the IEEE/CVF conference on computer vision and pattern recognition. pp. 22302--22313 (2023)

\bibitem{zhu2021perceptual}
Zhu, W., Zhai, G., Min, X., Yang, X., Zhang, X.P.: Perceptual quality assessment for recognizing true and pseudo 4k content. In: ICASSP 2021-2021 IEEE International Conference on Acoustics, Speech and Signal Processing (ICASSP). pp. 2190--2194. IEEE (2021)

\end{thebibliography}
\end{document}